\documentclass{article}

\usepackage[final,nonatbib]{neurips_2021}
\usepackage[utf8]{inputenc} % allow utf-8 input
\usepackage[T1]{fontenc}    % use 8-bit T1 fonts
\usepackage[dvipsnames]{xcolor}
\usepackage[pagebackref=true,breaklinks=true,letterpaper=true,colorlinks,bookmarks=false, citecolor=Blue]{hyperref}     % hyperlinks       %

\usepackage{url}            % simple URL typesetting
\usepackage{booktabs}       % professional-quality tables
\usepackage{amsfonts}       % blackboard math symbols
\usepackage{nicefrac}       % compact symbols for 1/2, etc.
\usepackage{microtype}      % microtypography
\usepackage{xcolor}         % colors
\usepackage{color}

\usepackage{blindtext}
\usepackage{enumitem}
\usepackage{float}
\usepackage{commath}
\usepackage{authblk}
\usepackage{csquotes}
\usepackage{amsmath,amssymb,amsthm}
\usepackage{hyperref}
\usepackage{graphicx}
\graphicspath{{./images/}{../images/} }
\usepackage{subfiles}
\usepackage{color}
\usepackage{diagbox}
\usepackage{multirow}
\usepackage{hhline}
\usepackage{xspace}
\usepackage{algorithm}
\usepackage{listings}

\renewcommand{\epsilon}{\varepsilon}
\renewcommand{\phi}{\varphi}

\newcommand{\BK}[1]{ {\left( #1 \right)} }
\newcommand{\sqBK}[1]{ {\left[ #1 \right]} }

\newcommand{\PP}{\ensuremath{\mathbb{P}}}

\newcommand{\EE}{\ensuremath{\mathbb{E}}}

\newcommand{\bp}{\mathbf{p}}
\newcommand{\bq}{\mathbf{q}}

\newcommand{\bw}{\mathbf{w}}
\newcommand{\netw}{\mathbf{f}}
\newcommand{\softmax}{\sigma_{\mathrm{SM}}}
\newcommand{\ent}{\mathcal{H}}
\newcommand{\bR}{\mathbb{R}}

\newcommand{\calX}{\mathcal{X}}
\newcommand{\calY}{\mathcal{Y}}

\newcommand{\calD}{\mathcal{D}}

\newcommand{\ECE}{\text{ECE}}

\newcommand{\aug}{\text{Aug}}
\newcommand{\train}{\text{train}}

\newcommand{\val}{\text{val}}

\newcommand{\avg}{\text{avg}}
\newcommand{\med}{\text{med}}
\newcommand{\trim}{\text{trim}}
\newcommand{\agg}{\textbf{Agg}}
\newcommand{\DC}{\text{DC}}

\DeclareMathOperator{\acc}{acc}
\DeclareMathOperator{\conf}{conf}
\DeclareMathOperator{\scale}{Scale}
\DeclareMathOperator{\ind}{\mathbf{1}}
\DeclareMathOperator*{\argmax}{arg\,max}
\DeclareMathOperator*{\argmin}{arg\,min}

\title{Uncertainty Quantification and Deep Ensembles}

\author{
Rahul Rahaman\\
Department of Statistics and Data Science,\\
National University of Singapore\\
\texttt{rahul.rahaman@u.nus.edu}
\and
Alexandre H. Thiery\\
Department of Statistics and Data Science,\\
National University of Singapore\\
\texttt{a.h.thiery@nus.edu.sg}
}

\begin{document}
\maketitle

\begin{abstract}
Deep Learning methods are known to suffer from calibration issues: they typically produce over-confident estimates. These problems are exacerbated in the low data regime.
Although the calibration of probabilistic models is well studied, calibrating extremely over-parametrized models in the low-data regime presents unique challenges.
We show that deep-ensembles do not necessarily lead to improved calibration properties.
In fact, we show that standard ensembling methods, when used in conjunction with modern techniques such as mixup regularization, can lead to less calibrated models.
This text examines the interplay between three of the most simple and commonly used approaches to leverage deep learning when data is scarce: data-augmentation, ensembling, and post-processing calibration methods. Although standard ensembling techniques certainly help boost accuracy, we demonstrate that the calibration of deep ensembles relies on subtle trade-offs. 
We also find that calibration methods such as temperature scaling need to be slightly tweaked when used with deep-ensembles and, crucially, need to be executed {\it after} the averaging process. Our simulations indicate that this simple strategy can halve the Expected Calibration Error (ECE) on a range of benchmark classification problems compared to standard deep-ensembles in the low data regime. Repository at: \href{https://github.com/rahulrahaman/Uncertainty-Quantification-and-Deep-Ensemble}{https://github.com/rahulrahaman/Uncertainty-Quantification-and-Deep-Ensemble}
\end{abstract}

\section{Introduction}
\label{sec.introduction}

%The high representational capacity of extremely over-parametrized neural networks gives them the ability to model datasets of ever-increasing complexity. Indeed, it has been reported that this class of models 
Overparametrized deep models can memorize datasets with labels entirely randomized~\cite{generalizationDeepLearning}. It is consequently not entirely clear why such extremely flexible models are able to generalize well on unseen data and trained with algorithms as simple as stochastic gradient descent, although a lot of progress on these questions have recently been reported~\cite{dziugaite2017computing,jacot2018neural,bietti2019inductive,mei2018mean,rotskoff2018neural,gabrie2018entropy}.

The high capacity of neural network models, and their ability to easily overfit complex datasets, makes them especially vulnerable to calibration issues. In many situations, standard deep-learning approaches are known to produce probabilistic forecasts that are over-confident~\cite{Guo_calibration_2017}. In this text, we consider the regime where the size of the training sets is very small, which typically amplifies these issues.
This can lead to problematic behaviors when deep neural networks are deployed in scenarios where a proper quantification of the uncertainty is necessary. Indeed, a host of methods~\cite{DeepEnsemble,SWAG,srivastava2014dropout,gal2016dropout,prechelt1998early} have been proposed to mitigate these calibration issues, even though no gold standard has so far emerged.
Many different forms of regularization techniques~ \cite{dataAugmentationEffectiveness, generalizationDeepLearning, regularizationElasticNet} have been shown to reduce overfitting in deep neural networks. Importantly, practical implementations and approximations of Bayesian methodologies~\cite{SWAG,SVDKL,VariationalBNNBlundel,PracVarInf,VariationDLMatGauss,rezende2014stochastic,MacKay_1992b} have demonstrated their worth in several settings. However, some of these techniques are not entirely straightforward to implement in practice. Ensembling approaches such as {\it drop-outs}~\cite{gal2016dropout} have been widely adopted, largely due to their ease of implementation. Recently, \cite{ashukha2021pitfalls} provides a study on different ensembling techniques and describes pitfalls of certain metric for in-domain uncertainty quantification. Also subsequent to our work, several articles also studied the interaction between data-augmentation and calibration issues. Importantly, the CAMixup approach is proposed as a promising solution in \cite{subsequentBalaji}. Furthermore, \cite{subsequentXixin} analyzes the under-confidence of ensembles due to augmentations from a theoretical perspective.  In this text, we investigate the practical use of Deep-Ensembles~\cite{DeepEnsemble,bonab2017less,lee2015m,szegedy2015going,fort2019deep,Guo_calibration_2017}, a straightforward approach that leads to state-of-the-art performances in most regimes. Although deep-ensembles can be difficult to implement when training datasets are large (but calibration issues are less pronounced in this regime), the focus of this text is the data-scarce setting where the computational burden associated with deep-ensembles is not a significant problem. 

% One aspect of our study is the effectiveness of Data Augmentation. It has become an integral part of DNN training. Augmentation policies have become increasingly rich, especially in the field of vision-related tasks. Apart from the popular image augmentations we also focus our study on data-agnostic augmentation \textit{mixup} \cite{mixup_original}. Recent studies \cite{mixup_calibration} suggest using mixup as a tool to produce models that possess good calibration and UQ properties. However, we believe that the level of exploration required in the data manifold to achieve good calibration is still partially understood. 

% On the other side, we study the behavior of a specific class of model averaging for Deep Neural Networks, named Deep Ensembles \cite{DeepEnsemble}. \cite{trustModelUncert} shows Deep Ensemble to be the best among a variety of Bayesian, semi-Bayesian, and non-Bayesian model averaging techniques when it comes to capturing uncertainty in model predictions. Hence we naturally focus on this method to bring the best out of a low-data setting.

\textbf{Contributions:} 
We study the interaction between three of the most simple and widely used methods for adopting deep-learning to the low-data regime: ensembling, temperature scaling, and mixup data augmentation. 
\begin{itemize}
\item 
Despite the widely-held belief that model averaging improves calibration properties, we show that, in general, standard ensembling practices do not lead to better-calibrated models. Instead, we show that averaging the predictions of a set of neural networks generally leads to less confident predictions: that is generally only beneficial in the oft-encountered regime when each network is overconfident. %Given the increasingly common association of good calibration property with Bayesian methods, it is crucial to point out that under some settings, calibration properties might deteriorate due to averaging. 
Although our results are based on Deep Ensembles, our empirical analysis extends to any class of model averaging, including sampling-based Bayesian Deep Learning methods.
\item 
We empirically demonstrate that networks trained with the {\it mixup} data-augmentation scheme, a widespread practice in computer vision, are typically under-confident. Consequently, subtle interactions between ensembling techniques and modern data-augmentation pipelines have to be considered for proper uncertainty quantification. The typical distributional shift induced by the mixup data-augmentation strategy influences the calibration properties of the resulting trained neural networks. In these settings, a standard ensembling approach typically worsens the calibration issues.
\item 
Post-processing techniques such as {\it temperature scaling} are sometimes regarded as competing methods when comparing the performance of many modern model-averaging techniques. Instead, to mitigate the under-confidence of model averaging, temperature scaling should be used \textit{in conjunction with} deep-ensembling methods. More importantly, the order in which the aggregation and the calibration procedures are carried out greatly influences the resulting uncertainty quantification.  These findings lead us to formulate the straightforward {\it Pool-Then-Calibrate} strategy for post-processing deep-ensembles: {\bf (1)} in a first stage, separately train deep models {\bf (2)} in a second stage, fit a {\it single} temperature parameter by minimizing a proper scoring rule (eg. cross-entropy) on a validation set. In the low data regime, this simple procedure can halve the Expected Calibration Error (ECE) on a range of benchmark classification problems when compared to standard deep-ensembles. Although straightforward to implement, to the best of our knowledge this strategy has not been investigated in the literature prior to our work.
\end{itemize}

\section{Background}
\label{sec.background}
%We first give a brief introduction to the setup and metrics that are used throughout the work. 
Consider a classification task with $C \geq 2$ possible classes $\calY \equiv \{1, \ldots, C\}$. For a sample $x \in \calX$, the quantity $\bp(x) \in \Delta_C = \{ \bp \in \bR^C_+ \, : \, p_1 + \ldots + p_C = 1\}$ represents a probabilistic prediction, often obtained as $\bp(x) = \softmax[\netw_{\bw}(x)]$ for a neural network $\netw_{\bw}: \calX \to \bR^C$ with weight $\bw \in \bR^D$ and softmax function $\softmax: \bR^C \to \Delta_C$. We set $\widehat{y}(x) \equiv \argmax_c p_c(x)$ and $\widehat{p}(x) = \max \bp(x)$.

% A model predicts class probabilities $\bp(x)$ for each observation $x \in \calX$. Note that $\bp(x)$ belongs to the probability simplex $\Delta_C = \{ \bp \in \bR^C_+ \, : \, p_1 + \ldots + p_C = 1\}$. A rather widely used approach is to first get a non-probabilistic output $\netw_{\bw}(x)$ from the model where $\netw_{\bw}: \calX \to \bR^C$ parameterized by model parameters $\bw$. Finally, the class probabilities are obtained by $\bp(x) = \softmax(\netw_{\bw}(x))$ where $\softmax: \bR^C \to \Delta_C$ is the softmax function,
% %
% \begin{align}
% \softmax(\bz)_i \; = \;  \frac{ \exp(z_i) }{\exp(z_1) + \ldots + \exp(z_C)}.
% \end{align}
% %
% The predicted class is chosen to be $\widehat{y} \equiv \argmax \bp ;\, \widehat{y} \in \calY \equiv \{1, \ldots, C\}$ with $\widehat{p}(x) = \max \bp(x)$ as the corresponding probability.

% \subsection{Augmentation}
\textbf{Augmentation:}
Consider a training dataset $\calD \equiv \{x_i, y_i\}_{i=1}^N$ and denote by $\overline{y} \in \Delta_C$ the one-hot encoded version of the label $y \in \calY$. A stochastic augmentation process $\aug: \calX \times \Delta_C \to \calX \times \Delta_C$ maps a pair $(x, \overline{y}) \in \calX \times \Delta_C$ to another augmented pair $(x_\star, \overline{y}_\star)$. In computer vision, standard augmentation strategies include rotations, translations, brightness and contrast manipulations. In this text, in addition to these standard agumentations, we also make use of the more recently proposed {\it mixup} augmentation strategy~\cite{mixup_original} that has proven beneficial in several settings. For a pair $(x,\overline{y}) \in \calX \times \Delta_C$, its mixup-augmented version $(x_{\star},\overline{y}_{\star})$ is defined as
\begin{align*}
x_\star = \gamma \, x + (1-\gamma) \, x_J
\quad \textrm{and} \quad 
\overline{y}_\star = \gamma \, \overline{y} + (1-\gamma) \, \overline{y}_J
\end{align*}
for a random coefficient $\gamma \in (0,1)$ drawn from a fixed mixing distribution often chosen as $\text{Beta}(\alpha, \alpha)$, and a random index $J$ drawn uniformly within $\{1, \ldots, N\}$. 

\textbf{Model averaging:}
Ensembling methods leverage a set of models by combining them into an aggregated model. In the context of deep learning, Bayesian averaging consists of weighting the predictions according to the Bayesian posterior $\pi(d \bw \mid \calD_{\train})$ on the neural weights. Instead of finding an optimal set of weights by minimizing a loss function, predictions are averaged. Denoting by $\bp_\bw(x) \in \Delta_C$ the probabilistic prediction associated to sample $x \in \calX$ and neural weight $\bw$, the Bayesian approach advocates to consider
\begin{align} \label{eq.bayes}
\textrm{(prediction)} \; \equiv \;
\int \bp_{\bw}(x) \, \pi(d \bw \mid \calD_{\train}) \in \Delta_C.
\end{align}
Designing sensible prior distributions is still an active area of research, and data-augmentation schemes, crucial in practice, are not entirely straightforward to fit into this framework. Furthermore, the high-dimensional integral \eqref{eq.bayes} is (extremely) intractable: the posterior distribution $\pi(d \bw | \calD_{\train})$ is  multi-modal, high-dimensional, concentrated along low-dimensional structures, and any local exploration algorithm (eg. MCMC, Langevin dynamics and their variations) is bound to only explore a tiny fraction of the state space. Because of the typically large number of degrees of symmetries, many of these local modes correspond to essentially similar predictions, indicating that it is likely not necessary to explore all the modes in order to approximate \eqref{eq.bayes} well. A detailed understanding of the geometric properties of the posterior distribution in Bayesian neural networks is still lacking, although a lot of recent progress has been made. Indeed, variational approximations have been reported to improve, in some settings, over standard empirical risk minimization procedures. Deep-ensembles can be understood as crude, but practical, approximations of the integral in Equation \eqref{eq.bayes}. The high-dimensional integral can be approximated by a simple non-weighted average over several modes $\bw_1, \ldots, \bw_K$ of the posterior distribution found by minimizing the negative log-posterior, or some approximations of it, with standard optimization techniques:
\begin{align*} % \label{eq.deep.ensemble}
\textrm{(prediction)} \; \equiv \;
\frac{1}{K}\Big\{ 
\bp_{\bw_1}(x) +  
\ldots +
\bp_{\bw_K}(x) \Big\}
\in \Delta_C.
\end{align*}
We refer the interested reader to~\cite{neal2012bayesian,mackay2003information,wilson2020bayesian, VariationalBNNBlundel} for different perspectives on Bayesian neural networks. Although simple and not well understood, deep-ensembles have been shown to provide highly robust uncertainty quantification when compared to more sophisticated approaches ~\cite{DeepEnsemble,bonab2017less,lee2015m,szegedy2015going}.

\textbf{Post-processing Calibration Methods:} 
The article \cite{Guo_calibration_2017} proposes a class of post-processing calibration methods that extend the more standard {\it Platt Scaling} approach~\cite{Platt}.
%where a linear transformation is applied to the logarithm of the probabilistic outputs. 
\textit{Temperature Scaling}, the simplest of these methods, transforms the probabilistic outputs $\bp(x) \in \Delta_C$ into a tempered version $\scale[\bp(x), \tau] \in \Delta_C$ defined through the scaling function
%$\scale: \Delta_C \times \bR_+ \to \Delta_C$,
%
\begin{align}\label{eqn:temp_scale}
\scale(\bp, \tau) \equiv \softmax \BK{ \log \bp / \tau  }
= \frac1Z \, \BK{p_1^{1/\tau}, \ldots, p_C^{1/\tau}} \in \Delta_C,
\end{align}
for a temperature parameter $\tau > 0$ and normalization $Z>0$.
The optimal parameter $\tau_\star > 0$ is usually found by minimizing proper-scoring rules~\cite{gneiting2007strictly}, often chosen as the negative log-likelihood, on a validation dataset.
Crucially, during this post-processing step, the parameters of the probabilistic model are kept fixed: the only parameter being optimized is the temperature $\tau>0$. 
In the low-data regime, the validation set being also extremely small, we have empirically observed that the more sophisticated {\it Vector} and  {\it Matrix} scaling post-processing calibration methods~\cite{Guo_calibration_2017} do not offer any significant advantage over temperature scaling approach and in fact overfit the extremely small validation dataset as chosen by our setup.

\textbf{Calibration Metrics:} 
% \begin{itemize}
%     \item  evaluation measure for probabilistic forecasts:
%     \item Brier score dates back to the 50s \cite{brier1950verification}
%     \item Proper scoring rules \cite{degroot1983comparison}
% \end{itemize}
%\red{** discuss reliability curves somewhere**} 
% Next we briefly describe some of the calibration metrics used throughout this work. One of the most commonly used and popular calibration metric \textit{Expected Calibration Error} or ECE. 
%Consider a probabilistic classifier $\bp:\calX \to \Delta_C$. 
The {\it Expected Calibration Error} ($\ECE$) measures the discrepancy between prediction confidence and empirical accuracy. 
% In this text, we also define the {\it signed Expected Calibration Error} ($\sECE$) in order to differentiate under-confidence from over-confidence. 
For a partition $0=c_0 < \ldots < c_M=1$ of the unit interval and a labelled set $\{x_i, y_i\}_{i=1}^N$, set $B_m = \{i : c_{m-1} < \widehat{p}(x_i) \leq c_m\}$. The quantity $\ECE$ is then defined as
\begin{gather} \label{eq.def.ECEs}
% \left\{
% \begin{aligned}
\ECE = \sum_{m=1}^{M}\frac{|B_m|}{N}\big|\conf_m - \acc_m \big|\\
\text{where}\quad \acc_m = \frac{1}{|B_m|}\sum_{i \in B_m}\ind(\widehat{y}(x_i) = y_i)
\quad \textrm{and} \quad
\conf_m = \frac{1}{|B_m|}\sum_{i \in B_m} \widehat{p}(x_i).
% \sECE = \sum_{m=1}^{M}\frac{|B_m|}{N}\big(\conf_m - \acc_m\big).
% \end{aligned}
% \right.
\end{gather}
A model is calibrated if $\acc_m \approx \conf_m$ for all $1 \leq m \leq M$. It is often instructive to display the associated {\it reliability curve}, i.e. the curve with $\conf_m$ on the $\text{x}$-axis and the difference $(\acc_m - \conf_m)$ on the $\text{y}$-axis. Figure \ref{image:pool_underconfident_1} displays examples of such reliability curves. A perfectly calibrated model is flat (i.e. $\acc_m - \conf_m=0$), while the reliability curve associated to an under-confident (resp. over-confident) model prominently lies above (resp. below) the flat line $\acc_m - \conf_m=0$.
We sometimes also report the value of the Brier score~\cite{brier1950verification} defined as $\frac{1}{N} \sum_{i=1}^{N} \|\bp(x_i) - \overline{y}_i \|_2^2$.

\textbf{Setup and implementation details:}
For our experiments, we use standard neural architectures. For CIFAR10/100 \cite{cifar10} we use ResNet18, ResNet34 \cite{resnet_paper} for Imagenette/Imagewoof \cite{imagenette-woof}, and for the  Diabetic Retinopathy  \cite{eyepacs_kaggle_DR}, similar to \cite{uncertaintyForDiabRetin} we use the architecture (not containing any residual connection) from the $5^{th}$ place solution of the associated {\it Kaggle} challenge. We also include the results for LeNet \cite{LeNet} trained on the MNIST \cite{lecun-mnisthandwrittendigit-2010} dataset in the appendix. A very low number of training examples (CIFAR$10\!: 1000,$ CIFAR$100\!: 5000,$ Image\{nette, woof\}: $5000$, MNIST: $500$) was used for all the datasets. However, we also show that our observations extend to full-data setups in \ref{sec.method}. The validation dataset is chosen from the leftover training dataset. The test dataset is kept as the original and is hidden during both training and validation step.
% Contrary to the suggested mixup augmentation level $(\alpha \le 0.4)$ in \cite{mixup_original}, we choose higher mixup $(\alpha \ge 0.8)$ levels because we observed improvement in performance in higher mixup across all our setups. We provide additional experiments supporting this in the appendix. 
%For example, with $1000$ training samples from the CIFAR10 dataset, a single ResNet18 model has $\approx 50\%$ test accuracy with $300$ epochs and without any augmentations. In contrast, the test accuracy reaches $\approx 70\%$ when usual data augmentation are used, the appropriate level of mixup, and deep ensemble is combined. 

%
%. === SECTION: EMPIRICAL OBSERVATION ====
%
\section{Empirical Observations}
\label{sec.empirical}
\begin{figure*}[ht]
\centering
\includegraphics[width=0.9\textwidth]{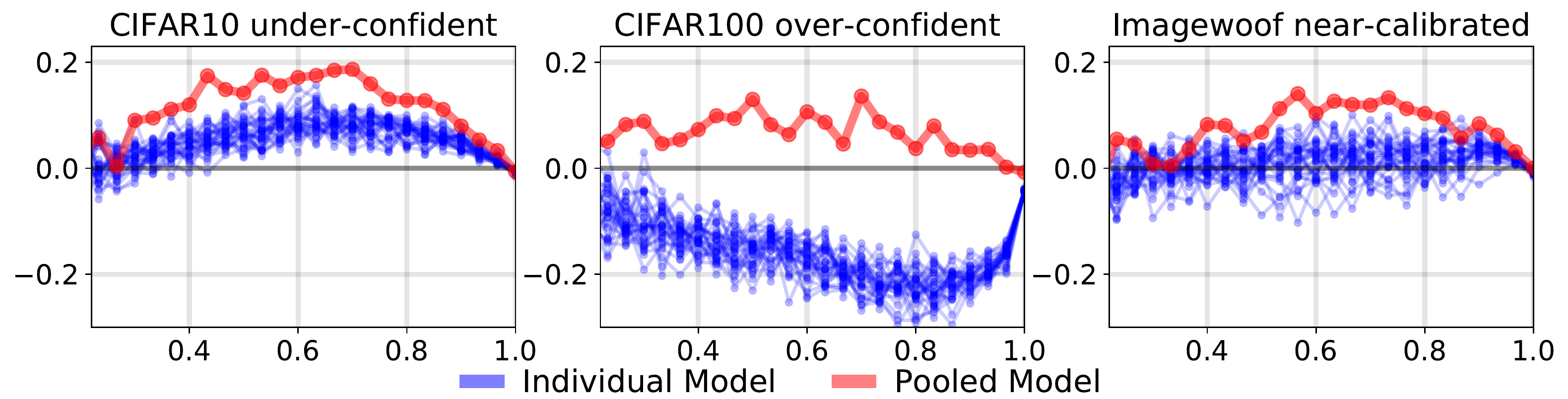}
\caption{\textbf{Confidence $\conf_m$ (x-axis) vs.\ Difference $(\acc_m - \conf_m)$ (y-axis):} We plot {\it Reliability Curves} in this figure, see Section \ref{sec.background} `Calibration Metrics' for definitions. The plots display the curves of $K=30$ individual networks \textbf{(blue)} trained on three datasets (i.e. CIFAR10, CIFAR100 and Imagewoof), as well as the pooled estimates \textbf{(red)} obtained by averaging the $K$ individual predictions. This linear averaging leads to consistently less confident predictions (i.e. higer values of $(\acc_m - \conf_m)$). It is only beneficial to calibration when each network is over-confident. It is typically detrimental to calibration when the individual networks are already calibrated, or under-confident.}
\label{image:pool_underconfident_1}
\end{figure*}

% In this section we gather several empirical facts related to the calibration properties of deep-neural networks when fitted on small datasets.
%
%
{\bf Linear pooling:} It has been observed in several studies that averaging the probabilistic predictions of a set of independently trained neural networks, i.e., deep-ensembles, often leads to more accurate and better-calibrated forecasts~\cite{DeepEnsemble,bonab2017less,lee2015m,szegedy2015going,fort2019deep}. Figure \ref{image:pool_underconfident_1} displays the reliability curves across three different datasets of a set of $K=30$ independently trained neural networks, as well as the reliability curves of the aggregated forecasts obtained by simply linear averaging the $K=30$ individual probabilistic predictions. These results suggest that deep-ensembles consistently lead to predictions that are {\it less confident} than the ones of its individual constituents. This can indeed be beneficial in the often encountered situation when each individual neural network is overconfident. Nevertheless, this phenomenon should not be mistaken with an intrinsic property of deep ensembles to lead to better-calibrated forecasts. For example, and as discussed further in Section \ref{sec.method}, networks trained with the popular {\it mixup} data-augmentation are typically under-confident. Ensembling such a set of individual networks typically leads to predictions that are even more under-confident.

\textbf{Other BNN methods:} It is important to point out that under-confidence of pooled predictions are not limited to Deep Ensembles. Other modern Bayesian Neural Network methods show similar properties. In table \ref{tab:other-BNN-method} we can see that ensembles obtained by SWAG \cite{SWAG} and MC-Dropout \cite{MCD}, two other popular model averaging techniques, are more under-confident than the individual models.

\begin{table}[ht]
    \centering
    \begin{tabular}{c|c c c}
        \hline
        \hline
         Dataset & Method & Single models & Ensemble\\
        \hline
        \multirow{2}{*}{CIFAR 10} & SWAG & 3.17 $\pm$ .27 & 4.36\\
         & MC-Dropout & 6.55 $\pm$ .10 & 7.59\\
        \hline
        \multirow{2}{*}{CIFAR 100} & SWAG & 3.34 $\pm$ .14 & 5.49\\
         & MC-Dropout & 4.92 $\pm$ .19 & 9.05\\
        \hline
        \hline
    \end{tabular}
    \caption{ECE, as defined in Equation \eqref{eq.def.ECEs}, of twenty individual models and the ensemble of SWAG \cite{SWAG} and MC-Dropout \cite{MCD} trained with mixup augmentation on full CIFAR\{10,100\} dataset. The ensemble is less calibrated than the individual models.}
    \label{tab:other-BNN-method}
\end{table}

In order to gain some insights into this phenomenon, recall the definition of the entropy functional $\ent: \Delta_C \to \bR$,
defined as $\ent(\bp) = - \sum_{k=1}^C p_k \, \log p_k$.
The entropy functional is concave on the probability simplex $\Delta_C$, i.e. $\ent(\lambda \bp + (1-\lambda) \, \bq) \geq \lambda \, \ent( \bp ) + (1-\lambda) \, \ent( \bq )$ for any $\bp, \bq \in \Delta_C$. Furthermore, tempering a probability distribution $\bp$ leads to an increased entropy if $\tau > 1$, as can be proved by examining the derivative of the function $\tau \mapsto \ent[\bp^{1/\tau}]$.
% , as can be seen from the relation
% %
% \begin{align}
% \frac{d}{d \tau} \ent[\scale(\bp, \tau)] \bigg\rvert_{\tau=1} 
% = 
% - \var_\bp \BK{ \text{U} } < 0
% \end{align}
% %
% with energy $\text{U}:\{1, \ldots, C\} \to \bR$ defined as $\text{U}: k \mapsto \log p_k$.
The entropy functional is consequently a natural surrogate measure of (lack of) confidence. The concavity property of the entropy functional shows that ensembling a set of $K$ individual networks leads to predictions whose entropies are higher than the average of the entropies of the individual predictions. In order to obtain a more quantitative understanding of this phenomenon, consider a binary classification framework. For a pair of random variables $(X,Y)$, with $X \in \calX$ and $Y \in \{-1,1\}$, and a classification rule $p: \calX \to [0,1]$ that approximates the conditional probability $p_x \approx \PP(Y=1 | X=x)$, define the {\it Deviation from Calibration} score as 
%
% == DC functional ===
%
\begin{align} \label{eq.def.DC}
\DC(p) \equiv \EE\sqBK{ \BK{\mathbf{1}_{\{Y=1\}} - p_X}^2  - p_X(1-p_X) }.
\end{align}
The term $\EE\sqBK{ \BK{\mathbf{1}_{\{Y=1\}} - p_X}^2}$ is equivalent to the Brier score of the classification rule $p$ and the quantity $\EE\sqBK{p_X(1-p_X)}$ is an entropic term (i.e. large for predictions close to uniform). Note that $\DC$ can take both positive and negative values and $\DC(p) = 0$ for a well-calibrated classification rule, i.e. $p_x = \PP(Y=1 | X=x)$ for all $x \in \calX$. Furthermore, among a set of classification rules with the same Brier score, the ones with less confident predictions (i.e. larger entropy) have a lesser $\DC$ score. In summary, the $\DC$ score is a measure of confidence that vanishes for well-calibrated classification rules, and that is low (resp. high) for under-confident (resp.over-confident) classification rules. Contrarily to the entropy functional, the $\DC$ score is extremely tractable. Algebraic manipulations readily shows that, for a set of $K \geq 2$ classification rules $p^{(1)}, \ldots, p^{(K)}$ and non-negative weights $\omega_1 + \ldots + \omega_K = 1$, the linearly averaged classification rule $ \sum_{i=1}^K \omega_i \, p^{(i)}$ satisfies
\begin{align} \label{eq.DC.decreaes}
\DC\BK{\sum_{i=1}^K \omega_i \, p^{(i)}}
\; = \;
\sum_{i=1}^K \omega_i \, \DC\BK{ p^{(i)}} 
\; - \; 
\underbrace{ \sum_{i,j=1}^{K} \, \omega_i \omega_j \, \EE\sqBK{ \BK{ p^{(i)}_X - p^{(j)}_X }^2} }_{ \geq 0}.
\end{align}

Equation \eqref{eq.DC.decreaes} shows that averaging classifications rules decreases the $\DC$ score (i.e. the aggregated estimates are less confident). Furthermore, the more dissimilar the individual classification rules, the larger the decrease. Even if each individual model is well-calibrated, i.e. $\DC(p^{(i)})=0$ for $1 \leq i \leq K$, the averaged model is not well-calibrated as soon as at least two of them are not identical.

\begin{figure*}[ht]
\centering
\includegraphics[width=1.0\textwidth]{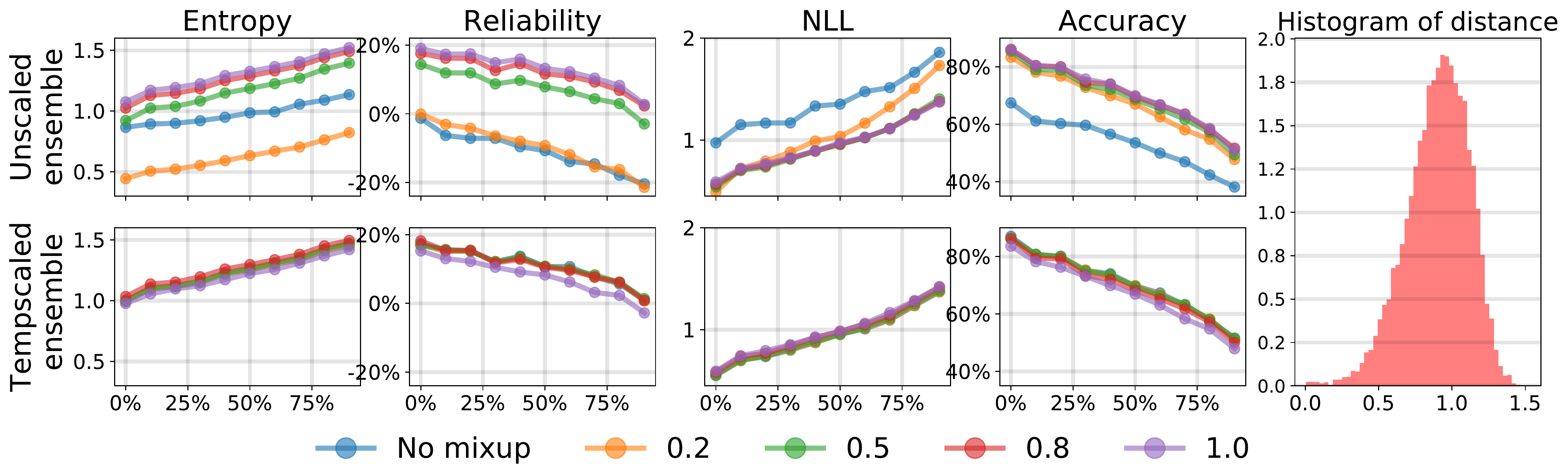}
\caption{\textbf{Metric (y-axis) vs. Distance from the training set (percentile) (x-axis)}: Deep Ensembles trained on $N=1000$ CIFAR10 samples with different amount of mixup regularization $\alpha \in \{.2, .5, .8, 1\}$. The $\text{x}$-axis represents quantiles of the distance to the CIFAR10 training set (see Section \ref{sec.empirical} for details). The overall distribution of the distances is displayed in the last column. The first row describes the performances of standard Deep Ensembles trained with data-augmentation and several amounts of mixup regularization $\alpha$. In the second row, before averaging the predictions of the members of the ensemble, each individual network is first temperature scaled on a validation set of size $N_{\text{val}}=50$: this corresponds to method {\bf (B)} of Section \ref{sec.method}.}
\centering
\label{image:distance-vs-metric}
\end{figure*}

\begin{figure*}[ht]
\centering
\includegraphics[width=1.0\textwidth]{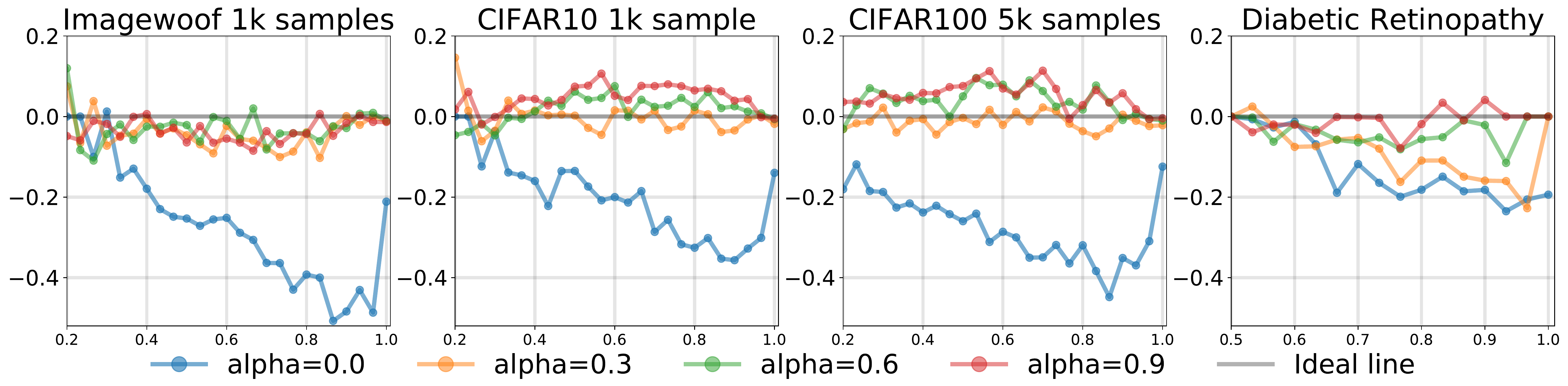}
\caption{ \textbf{Confidence $\conf_m$ (x-axis) vs.\ Difference $(\acc_m - \conf_m)$ (y-axis):} {\it Reliability curve} of a single neural network trained with different amount of mixup-augmentation on the Imagewoof, CIFAR10, CIFAR100 and Diabetic Retinopathy datasets. Increasing the amount of mixup augmentation, in general, makes the predictions {\it less}-confident. The case $\alpha=0$ corresponds to training without an mixup-augmentation, i.e. only using standard augmentation strategies.}
\label{image:alpha_and_calibration_1}
\end{figure*}

{\bf Distance to the training set:} In order to gain some additional insights into the calibration properties of neural networks trained on small datasets, as well as the influence of the popular mixup augmentation strategy, we examine several metrics (i.e., Accuracy, Reliability, Negative Log-likelihood (NLL), Entropy) as a function of the distance to the (small) training set $\calD_{\textrm{train}}$. The $2$nd column of Figure \ref{image:distance-vs-metric} displays the mean \textit{Reliability} (i.e., $\acc - \conf$) as a function of the distance percentiles. We focus on the CIFAR10 dataset and train our networks on a balanced subset of $N=1000$ training examples. Since there is no straightforward and semantically meaningful distance between images, we first use an unsupervised method (i.e., labels were not used) for learning a low-dimensional and semantically meaningful representation of dimension $d=128$. 
For these experiments, we obtained a mapping $\Phi: \bR^{32,32} \to \text{S}^{128}$, where $\text{S}^{128} \subset \bR^{128}$ denotes the unit sphere in $\bR^{128}$, with the {\it SimCLR} method \cite{chen2020simple}.
%, although experiments with other metric learning approaches~\cite{he2019momentum,ye2019unsupervised} have led to essentially similar conclusions. 
We used the distance $d(x,y) = \|\Phi(x) - \Phi(y)\|_2$, which in this case is equivalent to the cosine distance between the $128$-dimensional representations of the CIFAR10 images $x$ and $y$. The distance of a test image $x$ to the training dataset is defined as $\min\{d(x, y_i)\; : \; y_i \in \calD_{\textrm{train}}\}$. We computed the distances to the training set for each image contained in the standard CIFAR10 test set (last column of Figure \ref{image:distance-vs-metric}). Not surprisingly, we note that the average Entropy, Negative Log-likelihood, and Error Rate all increase for test samples further away from the training set. 
\begin{itemize}
\item {\bf Over-confidence:} The second column represents the \textit{Reliability curve}, but with bins (x-axis) as distance percentile, rather than confidence. The predictions associated with samples chosen further away from the training set have a lower value of $\acc - \conf$. This indicates that the {\it over}-confidence of the predictions increases (esp.\ lower mixup $\alpha$) with the distance to the training set. In other words, even if the entropy increases as the distance increases (as it should), calibration issues do not vanish as the distance to the training set increases. This phenomenon is irrespective of the amount of mixup used for training the network. 
\item {\bf Effect of mixup-augmentation:} The first row of Figure \ref{image:distance-vs-metric} shows that increasing the amount of mixup augmentation generally leads to an increase in entropy, decrease in over-confidence, as well as more accurate predictions (lower NLL and higher accuracy). Additionally, the effect is less pronounced for $\alpha \geq 0.2$. This is confirmed in Figure \ref{image:alpha_and_calibration_1} that displays more generally the effect of the mixup-augmentation on the reliability curves over four different datasets. In the appendix we provide more analysis on this.
% This is confirmed in Figure \ref{image:alpha_and_calibration_1} that displays the more generally the effect of the mixup-augmentation on the reliability curves over four different datasets.
%
\item {\bf Temperature Scaling:} Importantly, the second row of Figure \ref{image:distance-vs-metric} indicates that a post-processing temperature scaling for the individual models almost washes-out all the differences due to the mixup-augmentation scheme. For this experiment, an ensemble of $K=30$ networks is considered: before averaging the predictions, each network has been individually temperature scaled by fitting a temperature parameter (through negative likelihood minimization) on a validation set of size $N_{\text{valid}}=50$.
\end{itemize}

%=========================

\section{Calibrating Deep Ensembles}
\label{sec.method}
%\red{\bf calibration then pooling is bad }\\
% As discussed in the previous section, directly averaging a set of predictions leads to forecasts that are typically less confident than the individual models being averaged; it can consequently lead to an under-confident model. In practical terms, this means that the strategy consisting in calibrating each individual model before averaging them is likely to be sub-optimal. 

%
%. === POOLING MECHANISM ===
%
In order to calibrate deep ensembles, several methodologies can be considered: 

{\bf (A)} Do nothing and hope that the averaging process intrinsically leads to better calibration

{\bf (B)} Calibrate each individual network before aggregating all the results 

{\bf (C)} Simultaneously aggregate and calibrate the probabilistic forecasts of each individual model.

{\bf (D)} Aggregate first the estimates of each individual model before calibrating the pooled estimate.

% \begin{itemize}
% \item [{\bf (A)}] Do nothing and hope that the averaging process intrinsically leads to better calibration
% \item [{\bf (B)}] Calibrate each individual network before aggregating all the results 
% \item [{\bf (C)}] Simultaneously aggregate and calibrate the probabilistic forecasts of each individual model.
% \item [{\bf (D)}] Aggregate first the estimates of each individual model before eventually calibrating the pooled estimate.
% \end{itemize}
Simple pooling/aggregation rules that do not require a large number of tuning parameters are usually preferred, especially when training data is scarce \cite{jose2008simple,ProbForcastResearchPers}. Such rules are usually robust, conceptually easy to understand, and straightforward to implement and optimize. The standard and most commonly used average pooling of a set $\bp^{1:K}$ of $K \geq 2$ probabilistic predictions $\bp^{(1)}, \ldots, \bp^{(K)} \in \Delta_C \subset \bR^C$ is defined as
%
% \begin{align}
% \agg_{\avg}(\bp^{1:K}) = \frac{\bp^1 + \ldots + \bp^K}{K}
% \quad \textrm{and} \quad
% \agg_{\med}(\bp^{1:K}) = \frac{ \mathbf{median}(\bp^1, \ldots, \bp^K) }{\text{Z}},
% \end{align}
%
% \begin{align}
% \agg_{\avg}(\bp^{1:K}) &= \frac{\bp^1 + \ldots + \bp^K}{K}\\
% \agg_{\med}(\bp^{1:K}) &= \frac{ \mathbf{median}(\bp^1, \ldots, \bp^K) }{\text{Z}},
% \end{align}
\begin{align}\label{eqn:avg_pooling}
\agg(\bp^{1:K}) &= \frac{\bp^1 + \ldots + \bp^K}{K}.
\end{align}
Replacing the averaging with the median operation leads to \textit{median pooling} strategy, where the median is taken component-wise and then normalized afterward to obtain the final probability prediction. Alternatively, \textit{trimmed linear pooling} strategy removes a pre-defined percentage of outlier predictions before performing the average in \ref{eqn:avg_pooling}.
% for a normalization constant $\text{Z}>0$, the median operation being  executed component-wise over the $C \geq 2$ components. Finally, $\textbf{trim}(z^{1:K})$, the trimmed mean~\cite{jose2008simple} 
% %, also referred to as the truncated mean by some authors, 
% of $K$ real numbers $z_1, \ldots, z_K \in \bR$, is obtained by first discarding the $1 \leq \kappa \leq K /2$ largest and smallest values before averaging the remaining elements. This means that $\textbf{trim}(z^{1:K}) = [z_{\sigma(\kappa+1)} + \ldots z_{\sigma(K-\kappa-1)}] / (K-2\kappa)$ where $\sigma(\cdot)$ is a permutation such that $z_{\sigma(1)} \leq \ldots \leq z_{\sigma(K)}$. The trimmed mean pooling method is consequently defined as
% %
% \begin{align}
% \agg_{\trim}(\bp^{1:K}) = \frac{ \mathbf{trim}(\bp^1, \ldots, \bp^K) }{\text{Z}},
% \end{align}
% %
% for a normalization constant $\text{Z} > 0$, with the trimmed-averaging being executed component-wise. 
%

%
%. === AGGREGATION THEN CALIBRATION
%

{\bf Pool-Then-Calibrate (D):} any of the aforementioned aggregation procedure can be used as a pooling strategy before fitting a temperature $\tau_\star$ by minimizing proper scoring rules on a validation set. In all our experiments, we minimized the negative log-likelihood (i.e., cross-entropy). For a given set $\bp^{1:K}$ of $K \geq 2$ probabilistic forecasts, the final prediction is defined as 
\begin{align}
\bp_\star \equiv \scale\big[ \agg(\bp^{1:K}), \tau_\star \big]
\end{align}
where $\scale(\bp, \tau) \equiv \softmax \BK{ \log \bp / \tau }$.
Note that the aggregation procedure can be carried out entirely independently from the fitting of the optimal temperature $\tau_\star$.

\begin{figure*}[ht]
\centering
\includegraphics[width=0.9\textwidth]{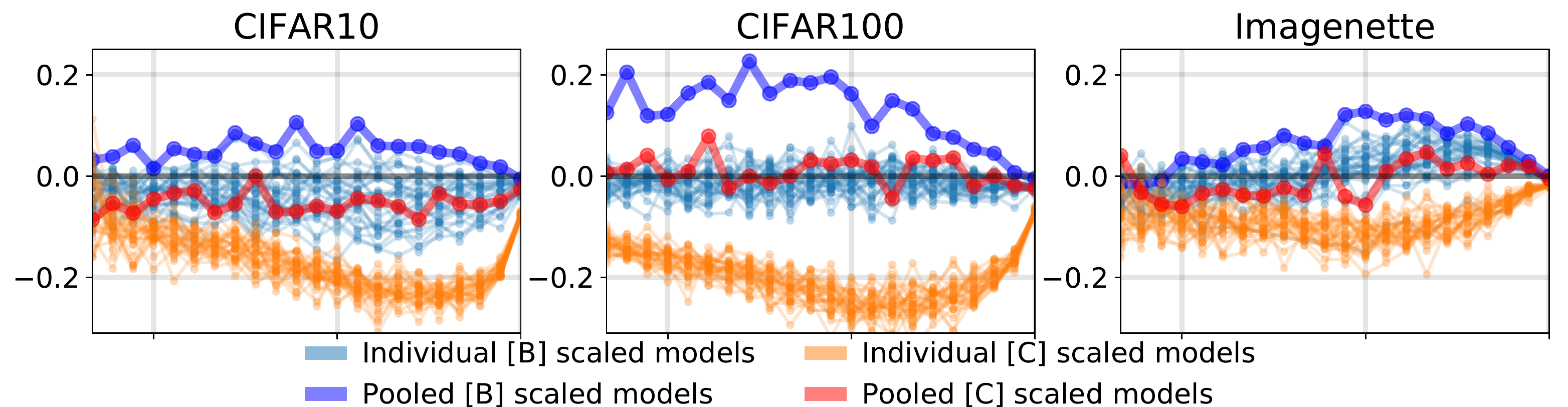}
\caption{\textbf{Confidence $\conf_m$ (x-axis) vs.\ Difference $(\acc_m - \conf_m)$ (y-axis):} Reliability curve of: 
{\bf (light blue)} each model calibrated with one temperature per model (i.e. individually temperature scaled),
{\bf (dark blue)} average of individually temperature scaled models (i.e. method {\bf [B]}),
{\bf (orange)} each model scaled with a global temperature obtained with method {\bf [C]},
{\bf (red)} result of method {\bf [C]} that consists in simultaneously aggregating and calibrating the probabilistic forecasts of each individual model.
{\bf Datasets:} a train:validation split of size $950:50$ was used for the CIFAR10 and IMAGENETTE datasets, and of size $4700:300$ for the CIFAR100 dataset. To avoid clutter, we omit method \textbf{[D]} for its similarity with method \textbf{[C]} in terms of performance.
}
\centering
\label{image:temperatures group B,C}
\end{figure*}
%%
%
%
%

%
% == JOINT OPTIMIZATION ===
%
{\bf Joint Pool-and-Calibrate (C):} there are several situations when the so-called {\it end-to-end} training strategy consisting in jointly optimizing several component of a composite system leads to increased performances~\cite{mnih2015human,mirowski2016learning,graves2016hybrid}. In our setting, this means learning the optimal temperature $\tau_\star$ concurrently with the aggregation procedure. The optimal temperature $\tau_\star$ is found by minimizing a proper scoring rule $\text{Score}(\cdot)$ on a validation set $\calD_{\text{valid}} \equiv \{x_i, y_i\}_{i=1}^{N_{\text{val}}}$,
\begin{align}\label{eq.group_c_methods}
\tau_\star = \argmin\Big\{  \tau \; \mapsto \;  \frac{1}{\calD_{\text{valid}}} \sum_{i \in \calD_{\text{valid}}} 
\text{Score}\BK{\bp_i^{\tau}, y_i} \Big\},
\end{align}
where $\bp^\tau_i \; = \; \agg\big[ \scale(\bp^{1:K}(x_i), \tau) \big] \in \Delta_C$ denotes the aggregated probabilistic prediction for sample $x_i$. 
In all our experiments, we have found it computationally more efficient and robust to use a simple grid search for finding the optimal temperature; we used $n=100$ temperatures equally spaced on a logarithmic scale in between $\tau_{\text{min}} = 10^{-2}$ and $\tau_{\text{max}} = 10$.

{\bf Importance of the Pooling and Calibration order:} 
Figure \ref{image:temperatures group B,C} shows calibration curves when individual models are temperature scaled separately (i.e. group \textbf{[B]} of methods), as well as when the models are scaled with a common temperature parameter (i.e. group \textbf{[C]} of methods). Furthermore, the calibration curves of the pooled model (group \textbf{[B]} and \textbf{[C]} of methods) are also displayed. More formally, the group \textbf{[B]} of methods obtains for each individual model $1\!\leq\!k\!\leq\!K$ an optimal temperature $\tau_{\star}^{(k)} > 0$  as solution of the optimization procedure
\begin{equation*}
\tau_{\star}^{(k)} = \argmin_{\tau} \;  \frac{1}{\calD_{\text{valid}}} \sum_{i \in \calD_{\text{valid}}} 
\text{Score}\BK{\scale\big[\bp_i^{k}, \tau\big], y_i}
\end{equation*}
% \begin{equation*}
% \tau_{\star}^{(k)} = \argmin\bigg\{  \tau \; \mapsto \;  \frac{1}{\calD_{\text{valid}}} \sum_{i \in \calD_{\text{valid}}} 
% \text{Score}\BK{\scale\big[\bp_i^{k}, \tau\big], y_i} \bigg\}
% \end{equation*}
%
where $\bp_i^{k} \in \Delta_C$ denotes the probabilistic output of the $k^{th}$ model for the $i^{th}$ example in validation dataset. The \textit{light blue} calibration curves corresponds to the outputs $\scale\big[\bp^{k}, \tau^{(k)}_{\star} \big]$ for $K$ different models. The \textit{deep blue} calibration curve corresponds the linear pooling of the individually scaled predictions. 
For the group \textbf{[C]} of methods, a single common temperature $\tau_\star > 0$ is obtained as solution of the optimization procedure described in equation \eqref{eq.group_c_methods}.
% \begin{align}
% \tau_\star = \argmin\bigg\{  \tau \; \mapsto \;  \frac{1}{\calD_{\text{valid}}} \sum_{i \in \calD_{\text{valid}}} 
% \text{Score}\BK{\bp_i^{\tau}, y_i} \bigg\},
% \end{align}
%
The \textit{orange} calibration curves are generated using the predictions $\scale\big[\bp^{k}, \tau_{\star}\big]$, and the \textit{red} curve corresponds to the prediction $\agg\big[ \scale(\bp^{1:K}, \tau_{\star}) \big]$.
Notice that when scaled separately (by $\tau^{(k)}_{\star}$), each of the individual models (light blue) is close to being calibrated, but the resulting pooled model (deep blue) is under-confident. However, when scaled by a common temperature, the optimization chooses a temperature $\tau_{\star}$ that makes the individual models (orange) slightly over-confident so that the resulting pooled model (red) is nearly calibrated. This reinforces the justifications in section \ref{sec.empirical}, and it also shows the importance of the order of pooling and scaling.

\begin{figure*}[ht]
\centering
\includegraphics[width=1.0\textwidth]{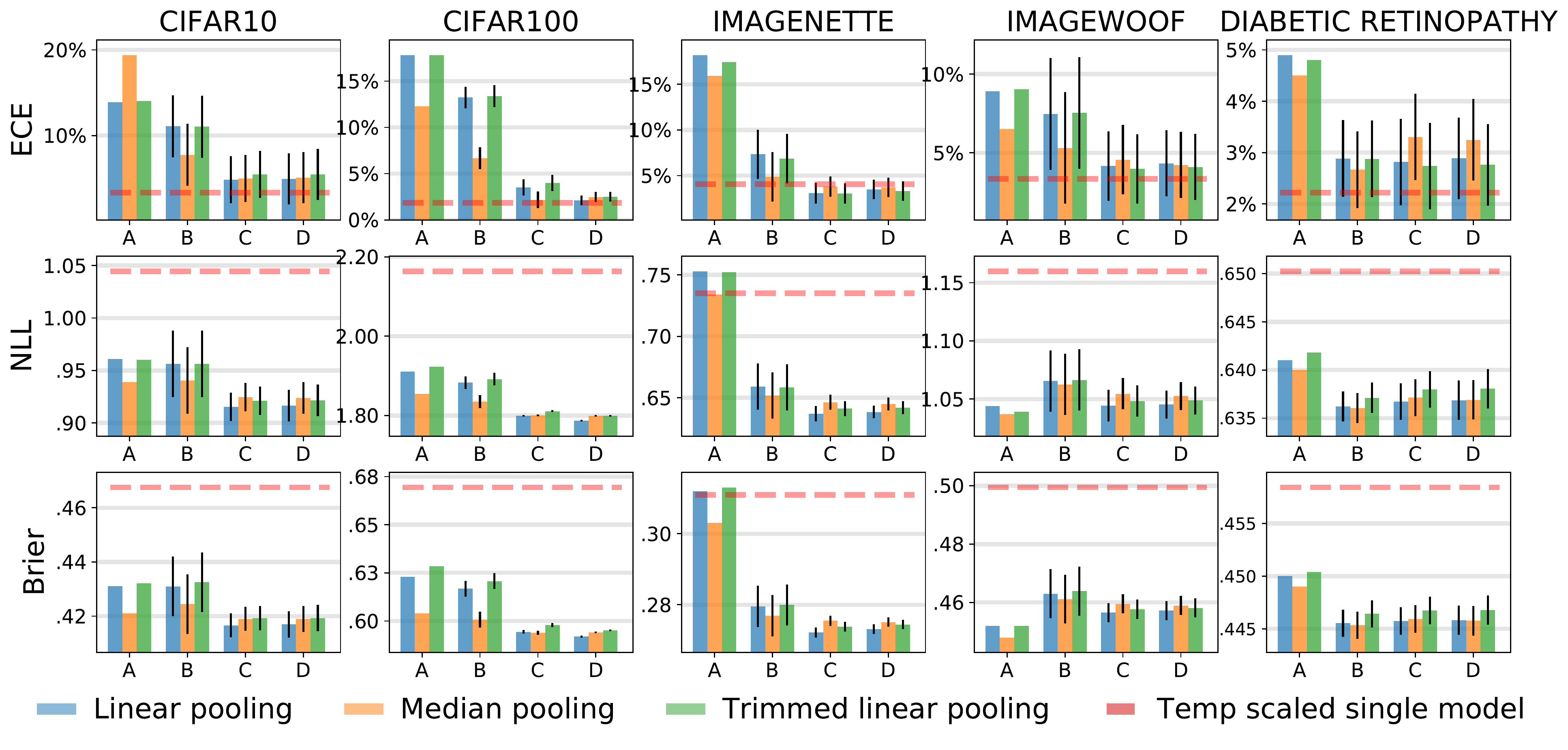}
\caption{\textbf{Pooling method (x-axis) vs.\ Metric (y-axis):} Performance of different pooling strategies (\textbf{A-D}) with $K=30$ models trained with mixup-augmentation ($\alpha=1$) across multiple datasets. The total datasets (training + validation) were of size $N=1000$ for CIFAR10 and Imagenette and Imagewoof, and $N=5000$ for CIFAR100 and Diabetic Retinopathy. Experiments were executed $50$ times on the same training data but different validation sets. The dashed red line represents a baseline performance when a single model was training with mixup augmentation ($\alpha = 1$) and post-processed with temperature scaling.}
\label{image:temp_scale_performance}
\end{figure*}

Figure \ref{image:temp_scale_performance} compares the four methodologies {\bf A-B-C-D} identified at the start of this section, with the three different pooling approaches $\agg_{\avg}$ and $\agg_{\med}$ and $\agg_{\trim}$. These methods are compared to the baseline approach (in dashed red line) consisting of fitting a single network trained with the same amount $\alpha=1$ of mixup augmentation before being temperature scaled. All the experiments are executed $50$ times, on the {\it same} training set, but with $50$ different validation sets of size $N_{\val} = 50$ for CIFAR10, Imagenette, Imagewoof and $N_{\val} = 300$ for CIFAR100, and $N_{\val} = 500$ for the Diabetic Retinopathy dataset. The results indicate that on most metrics and datasets, the (naive) method ${\bf (A)}$ consisting of simply averaging predictions is not competitive. Secondly, and as explained in the previous section, the method {\bf (B)} consisting in first calibrating the individual networks before pooling the predictions is less efficient across metrics than the last two methods ${\bf (C-D)}$. Finally, the two methods ${\bf (C-D)}$ perform comparably, the method {\bf (D)} (i.e. {\it pool-then-calibrate}) being slightly more straightforward to implement. With regards to the pooling methods, the intuitive robustness of the {\it median} and {\it trimmed-averaging} approaches do not seem to lead to any consistent gain across metrics and datasets. Note that ensembling a set of $K=30$ networks (without any form of post-processing) does lead to a very significant improvement in NLL and Brier score but leads to a serious deterioration of the ECE. The {\it Pool-Then-Calibrate} keeps the gains in NLL/Brier score unaffected, without compromising calibration.

{\bf Importance of the validation set:} it would be practically useful to be able to fit the temperature without relying on a validation set. We report that using the training set instead (obviously) does not lead to better-calibrated models. We have tried to use a different amount of mixup-augmentation (and other types of augmentation) on the training set for fitting the temperature parameter but have not been able to obtain satisfying results.

\begin{figure*}[ht]
\centering
\includegraphics[width=1.0\textwidth]{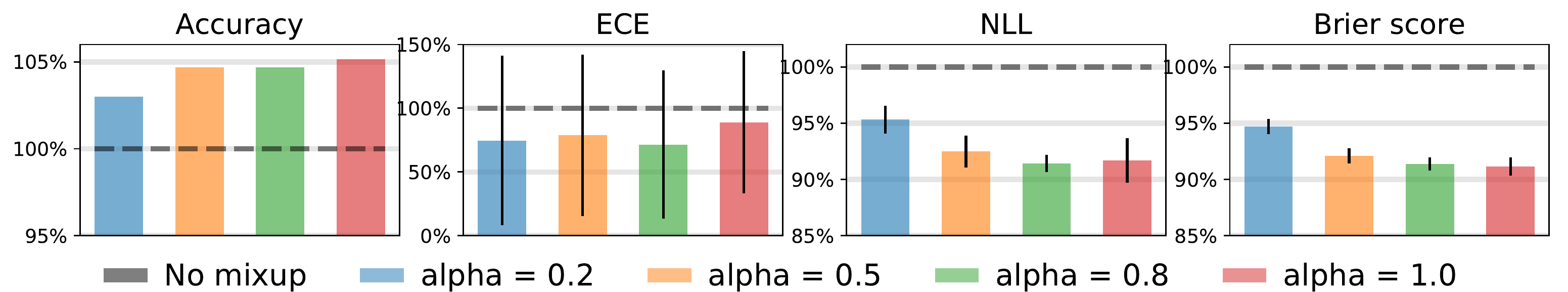}
\caption{ \textbf{Mixup strength (x-axis) vs.\ Metric (y-axis):} {\it Pool-Then-Calibrate} approach when applied to a deep-ensemble of $K=30$ networks trained with different amount of mixup-augmentation on $N=1000$ CIFAR10 training samples ($N_{\val} = 50$ were used for validation). 
For each metric, we report the {\bf ratio} of performance when compared to the {\it Pool-then-Calibrate} method used without any form of mixup-augmentation (only standard data-augmentation). The results indicate a clear benefit in using the mixup-augmentation in conjunction with temperature scaling. The error bars represent variability due to the choice of different validation sets.} 
% Experiments were executed on $50$ different validation sets (the errorbars show the variation), and a fixed training set of $950$ samples.}
\label{image:performance_gain}
\end{figure*}

{\bf Role and effect of mixup-augmentation:}
the mixup augmentation strategy is popular and straightforward to implement. As already empirically described in Section \ref{sec.empirical}, increasing the amount of mixup-augmentation typically leads to a decrease in the confidence and increase in entropy of the predictions. This can be beneficial in some situations but also indicates that this approach should certainly be employed with care for producing calibrated probabilistic predictions. 
Contrarily to other geometric data-augmentation transformations such as image flipping, rotations, and dilatations, the mixup strategy produces non-realistic images that consequently lie outside the data-manifold of natural images: leading to a large distributional shift. Mixup relies on a subtle trade-off between the increase in training data diversity, which can help mitigate over-fitting problems, and the distributional shift that can be detrimental to the calibration properties of the resulting method. 
Figure \ref{image:performance_gain} compares the performance of the {\it Pool-Then-Calibrate} approach when applied to a deep ensemble of $K=30$ networks trained with different amounts of mixup-$\alpha$. The results are compared to the same approach (i.e. {\it Pool-then-Calibrate} with $K=30$ networks) with no mixup-augmentation. The results indicate a clear benefit in using the mixup-augmentation in conjunction with temperature scaling.

{\bf Extension to full-data setting:}
Although classification accuracy is usually not an issue when data is plentiful, the lack of calibration can indeed be still present when models are trained with aggressive data-augmentation strategies (as is common nowadays): the distributional shift between (data-augmented) training samples and (non-augmented) test samples when models are used in production can lead to significant calibration issues. Although we mainly focus on low-data setting, below in table \ref{tab:full-data-setting} we show that our conclusion extends to full-data setting as well. We have investigated below the CIFAR100 full dataset (ResNet architecture / no-mixup) setting under varying conditions.

\begin{table}[ht]
    \centering
    \begin{tabular}{c|c c c c}
        \hline
        \hline
         Method	& Accuracy & ECE & NLL & Brier\\
        \hline
        (1) Individual models (unscaled) & $70.8 \pm .36$ & $9.8 \pm .31$ & $1.17 \pm.01$ & $0.411$\\
        Ensemble of models in (1) & $78.4$ & $5.9$ & $0.782$ & $0.308$\\
        \hline
        (2) Individual models (temp scaled)	& $70.8 \pm .36$ & $2.1 \pm .4$ & $1.07 \pm .01$ & $0.396$\\
        Ensemble of models in (2) & $78.4$ & $13.2$ & $0.859$ & $0.331$\\
        \hline
        Pool-then-calibrate & $78.4$ & $3.4$ & $0.770$ & $0.303$\\
        \hline
        \hline
    \end{tabular}
    \caption{In line with our discussion in Sec \ref{sec.empirical}, we show that linear pooling \textbf{(A)} appears to be helping with calibration ($2^{nd}$ row) when individual models are mildly over-confident ($1^{st}$ row), but performs worse ($4^{th}$ row) than individual models even in full-data setting (CIFAR100 50K training) when the individual models are near-calibrated ($3^{rd}$ row). Our proposed \textit{pool-then-calibrate} \textbf{(D)} has the best performance ($5^{th}$ row).}
    \label{tab:full-data-setting}
\end{table}

% Method	Accuracy	ECE	NLL	Brier
% (1) Individual models (unscaled)	70.8 ± .36	9.8% ± .31	1.17 ± .01	0.411 ± .0
% Ensemble of models in (1)	78.4 & 5.9	0.782	0.308
% (2) Individual models (temp scaled)	70.8% ± .36	2.1% ± .4	1.07 ± .01	0.396 ± .0
% Ensemble of models in (2)	78.4%	13.2%	0.859	0.331
% Pool-then-calibrate	78.4%	3.4%	0.770	0.303

The first row reports the performance of individual models trained without mixup: the individual models are over-confident, but not extremely over-confident (presumably because of the large number of samples). When these models are pooled to make an ensemble in the second row, the pooled model is better calibrated. This is the setup that is usually studied in almost every early articles investigating the properties of deep-ensembles, hence leading to the conclusion that deep-ensembling inherently brings calibration. When we make the individual models calibrated in the $3^{rd}$ row, where we used temp-scaling but it can also be due to the effect of more aggressive data-augmentation schemes, the individual calibration naturally improves significantly. Nevertheless, when we pool these calibrated models to make an ensemble, the pooled model suffers from extreme under-confidence ($4^{th}$ row). Our proposed method \textit{pool-then-calibrate} ($5^{th}$ row) performs well even in full-data setting.

\textbf{Out-of-distribution performance:} We show the out-of-distribution detection performance of our method compared to vanilla ensembling when the ensembles are trained on CIFAR10 and tested on a subset of CIFAR100 classes which are visually different from CIFAR10. In table \ref{tab:ood-experiment}, we show the metric: difference between the medians of the in-class and out-of-class prediction entropy (higher is better).

\begin{table}[ht]
    \centering
    \begin{tabular}{c|c | c}
        \hline
        \hline
        Single model 30 variations & Deep Ensemble [A] & Pool-then-calibrate [D]\\
        \hline
        $0.342 \pm 0.015$ & $0.359$ & $0.521$\\
        \hline
    \end{tabular}
    \caption{Difference in median prediction entropy between in-class (CIFAR10) vs out-of-class (CIFAR100 subset) dataset. Pool-then-calibrate brings significant improvement in terms of out-of-distribution detection.}
    \label{tab:ood-experiment}
\end{table}

Pool-then-Calibrate performs significantly better than vanilla ensemble in separating the predictions for in-class and out-of-class observations (45\% more separation in terms of distance between medians). In table \ref{tab:cifar-10c}, we also show the performance when we run inference on the CIFAR10-C dataset (Gaussian noise) after training our ensemble model on the setting: 1000 samples of CIFAR10 dataset with mixup 1.0. As expected, vanilla ensembling with linear pooling (\textbf{A}) has worse calibration than single models, while pool-then-calibrate (\textbf{D}) improves score across the board.

\begin{table}[ht]
    \centering
    \begin{tabular}{c|c c c c}
        \hline
        \hline
        Method & Accuracy & ECE & NLL & Brier\\
        \hline
        Individual models & $59.38 \pm 0.05$ & $6.57 \pm 0.006$ & $1.237 \pm 0.012$ & $0.549 \pm 0.005$\\
        Deep Ensemble \textbf{[A]} & $64.63$ & $15.13$ & $1.145$ & $0.511$\\
        Pool-then-Calibrate \textbf{[D]} & $64.63$ & $1.65$ & $1.059$ & $0.480$\\
        \hline
    \end{tabular}
    \caption{Inference on CIFAR-10C (Gaussian noise), trained on CIFAR10 (1K sample). Pool-then-calibrate \textbf{[D]} performs better while vanilla ensemble \textbf{[A]} has worse calibration than single models.}
    \label{tab:cifar-10c}
\end{table}

\textbf{Additional experiments: } In the appendix, we add more experiments on the effect of number of models in the ensemble, detailed numerical results for all datasets as well as MNIST, ablation study, and effect of different mixup levels on all the metrics.

{\bf Cold posteriors:} the article \cite{wenzel2020good} reports gains in several metrics when fitting Bayesian neural networks to a {\it tempered posterior} of type $\pi_\tau(\theta) \propto \pi(\theta)^{1/\tau}$, where $\pi(\theta)$ is the standard Bayesian posterior, for temperatures $\tau$ {\it smaller than one}. Although not identical to our setting, it should be noted that in all our experiments, the optimal temperature $\tau_\star$ was consistently smaller than one. In our setting, this is because simply averaging predictions lead to under-confident results. We postulate that related mechanisms are responsible for the observations reported in \cite{wenzel2020good}.

% \textbf{Additional experiments:}
% The {\it appendix Material} reports experiments that investigate the dependence of our results on the number $K \geq 2$ of models in the ensembles, out-of-sample detection, and justification for using higher mixup parameters $\alpha$. In addition, the {\it appendix Material} presents detailed numerical results and ablation studies, as well as  training-evaluation scripts to facilitate the replication of results.

%=========================
\section{Discussion}
\label{sec.discussion}
The problem of calibrating deep-ensembles has received surprisingly little attention in the literature. In this text, we examined the interaction between three of the most simple and widely used methods for adopting deep-learning to the low-data regime: ensembling, temperature scaling, and mixup data augmentation. We highlight that ensembling in itself does not lead to better-calibrated predictions, that the mixup augmentation strategy is practically important and relies on non-trivial trade-offs, and that these methods subtly interact with each other. Crucially, we demonstrate that the order in which the pooling and temperature scaling procedures are executed is important to obtaining calibrated deep-ensembles. We advocate the {\it Pool-Then-Calibrate} approach consisting of first pooling the individual neural network predictions together before eventually post-processing the result with a simple and robust temperature scaling step. 
% Furthermore, we note that this approach is insensitive to the choice of pooling method, the simple linear averaging procedure being essentially as robust as the others.

\section{Broader Impact}
Producing well-calibrated probabilistic predictions is crucial to risk management, and when decisions that rely on the outputs of probabilistic models have to be trusted. Furthermore, designing well-calibrated models is crucial to the adoption of machine-learning methods by the general public, especially in the field of AI-driven medical diagnosis, since it is intimately related to the issue of trust in new technologies. 

\appendix

\section{Additional experiments}

\begin{figure}[ht]
\includegraphics[width=\textwidth]{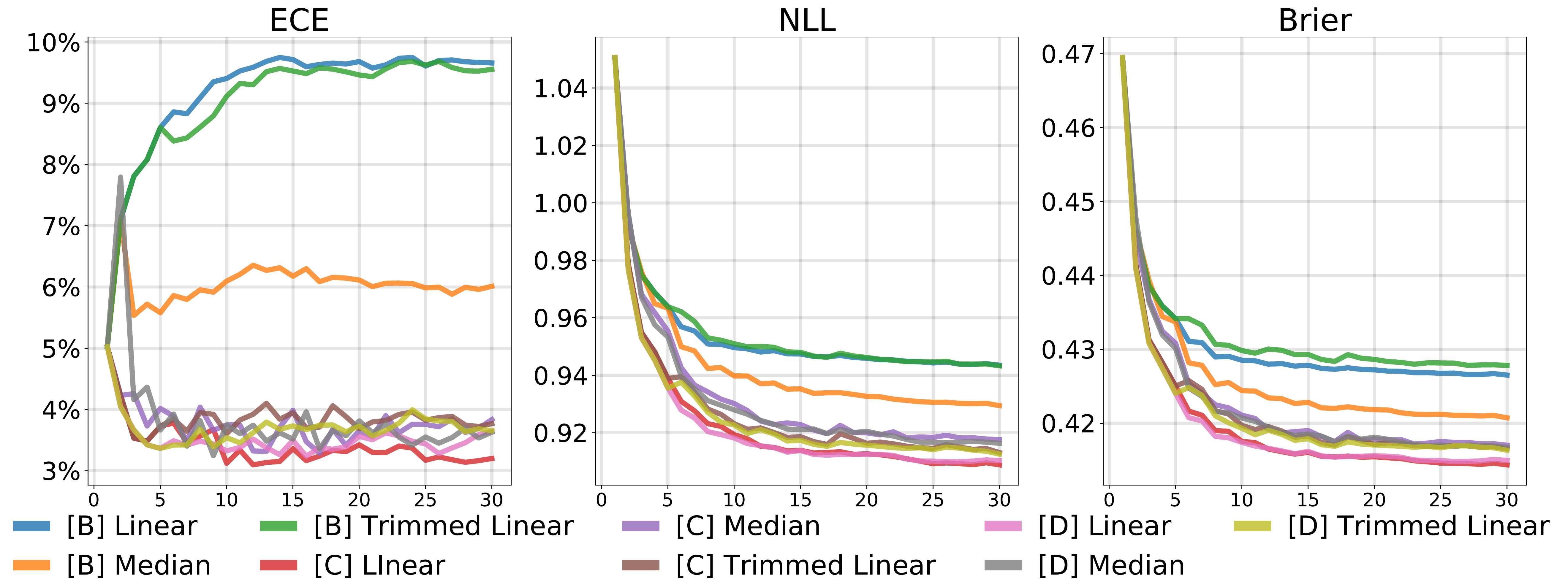}
\caption{Comparison of methods {\bf B-C-D} described at the start of Section \ref{sec.method} on the CIFAR10 dataset with $N=1000$ samples (950:50 split). The $x$-axis denotes the number of models. To avoid clutter and due to significantly worse performance, method {\bf [A]} (i.e. standard deep-ensemble without any form of calibration) is omitted.}
\centering
\label{image:num model vs performance}
\end{figure}

\paragraph{Size of the ensembles}
Figure \ref{image:num model vs performance} shows the performance of the different pooling methods (i.e. groups \textbf{[B]-[D]}) on the CIFAR10 dataset, as a function of the number of individual models in the ensemble. For clarity, the (non-calibrated) group \textbf{[A]} of methods are not reported. Recall that the group \textbf{[A]} pools the the predictions without any calibration procedure, the group \textbf{[B]} first calibrates each individual models separately before aggregating the results, the group \textbf{[C]} jointly calibrates and aggregates the prediction, and finally the group \textbf{[D]} first aggregates the results before calibrating the resulting prediction. Methods in group \textbf{[C]} and \textbf{[D]} performs similarly. For the CIFAR10 dataset, we observe that the performance under most metrics saturates for ensemble of sizes $\approx 15$.\\

\begin{table}[h]
    \centering
    \begin{tabular}{c|c|c|c|c|c|}
        \cline{2-6}
         & Num model & 1 & 4 & 8 & 15 \\
         \hline
         \multicolumn{1}{|c|}{Deep Ensemble} & ECE & $7.31$ & $12.37$ & $13.44$ & $13.87$\\
         \multicolumn{1}{|c|}{method \textbf{[A]}} & Brier & $0.464$ & $0.440$ & $0.435$ & $0.432$ \\
         \hline
         \multicolumn{1}{|c|}{\textit{Pool-then-calibrate}} & ECE & -- & $3.44$ & $2.99$ & $3.17$\\
         \multicolumn{1}{|c|}{method \textbf{(D)}} & Brier & -- & $0.415$ & $0.410$ & $0.406$ \\
         \hline
    \end{tabular}
    \caption{CIFAR10: Influence of the Ensemble Size}
    \label{tab:varying_models}
\end{table}

\begin{figure}[h]
    \includegraphics[width=\textwidth]{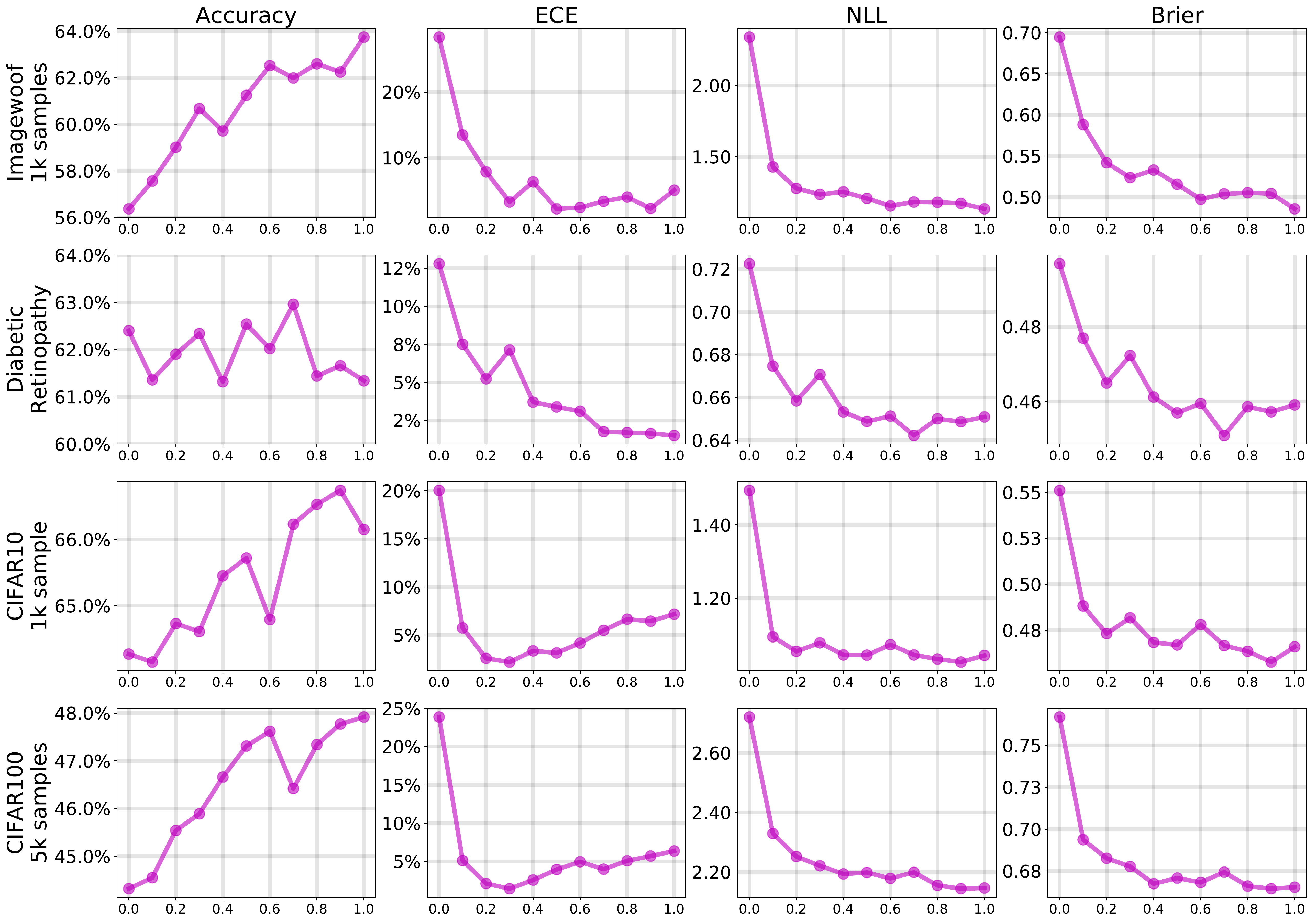}
        \caption{\textbf{Mixup $\alpha$ (x-axis) vs.\ Metric (y-axis)}: The effect of a higher mixup in NLL, ECE, BRIER score is quite evident in the plots. In our setting, most of the metrics improve as a function of $\alpha$. The CIFAR\{10,100\} datasets show a slight increment in the ECE because the model starts to become under-confident. In contrast, the other three metrics for CIFAR show improvement. From top to bottom, the datasets are Imagewoof 1000 samples, Diabetic Retinopathy with 5000 samples, CIFAR10 with 1000 samples, and CIFAR100 with 5000 samples. The metric in each row is test accuracy, test ECE, test NLL, and test Brier from left to right.}
    \centering
    \label{image:highmixup_performance_alldset}
\end{figure}
\paragraph{Effect of mixup $\alpha$}
In figure \ref{image:highmixup_performance_alldset} we list generalization and calibration results of high $\alpha$ mixup augmentation. All the setups in which we analyze the performance are limited in the number of training data points. It shows that even if with adequate data, high mixup makes models under-confident; for low data settings, mixup with $\alpha$ near 1.0 boosts model performance quite significantly.

\begin{table}[h]
    \centering
    \begin{tabular}{|c|c|c|c|c|c|}
    \hline
        \multirow{3}{*}{Metric} & (Ours) 30 models & 30 models & single model & single model & single model\\
        & Method \textbf{[D]} & mixup & mixup & no mixup & no mixup\\
        & Augment + mixup & Augment & Augment & Augment & no Augment\\
        \hline
        test acc & 69.92 $\pm$ .04 & \textbf{70.67} & 66.45 $\pm$ .61 & 63.73 $\pm$ .51 & 49.85 $\pm$ .66 \\
        test ECE & \textbf{3.3} $\pm$ 1.9 & 13.9 & 7.03 $\pm$ .7 & 20.7 $\pm$ .4 & 23.4 $\pm$ 1.0 \\
        test NLL & \textbf{0.910} $\pm$ .012 & 0.961 & 1.03 $\pm$ .13 & 1.509 $\pm$ .017 & 1.770 $\pm$ .045 \\
        test BRIER & \textbf{0.414} $\pm$ .002 & 0.431 & 0.463 $\pm$ .005 & 0.556 $\pm$ .006 & 0.718 $\pm$ .009 \\
    \hline
    \end{tabular}
    \caption{Ablation study performed on CIFAR10 1000 samples. For ensemble temp scaling, we use 950 training samples and 50 validation sets. For setups with variation, we report metric mean and standard deviation.}
    \label{tab:ablation}
\end{table}

\paragraph{Ablation study:} We focus on the CIFAR10 dataset with $N_{train} = 1000$ fixed training examples, and $100$ different validation sets of size $N_{val} = 50$: Table \ref{tab:ablation} reports the means and standard deviations across these experiments. For setups involving training a single model, we report the mean and standard deviations of the metric from a variety of 30 different trained models.

\begin{table}[h]
    \centering
    \begin{tabular}{|c|c|c|c|c|}
        \multicolumn{5}{c}{CIFAR10 - 1000 samples}\\
        \hline
        Method & Test Accuracy & Test ECE & Test NLL & Test Brier\\
        \hline
        Single model & 66.48 $\pm$ .62 & 7.31 $\pm$ .7 & 1.037 $\pm$ .013 & 0.464 $\pm$ .005\\
        Vanilla pooling \textbf{[A]} & 70.71 & 13.9 & 0.961 & 0.431\\
        Pool-then-calibrate \textbf{[D]} & 70.71 & 4.9 $\pm$ 2.9 & 0.916 $\pm$ .015 & 0.417 $\pm$ .005 \\
        \hline
        
        \multicolumn{5}{c}{CIFAR100 - 5000 samples}\\
        \hline
        Single model & 46.8 $\pm$ .41 &  5.4 $\pm$ .37 &  2.180 $\pm$  0.014 &  0.674 $\pm$  0.003\\
        Vanilla pooling \textbf{[A]} & 55.32 & 17.8 & 1.911 & 0.623\\
        Pool-then-calibrate \textbf{[D]} & 55.32 & 2.1 $\pm$ .5 & 1.787 $\pm$ .002 & 0.592 $\pm$ .0\\
        \hline
        
        % \multicolumn{5}{c}{CIFAR100 - full dataset (10000 samples)}\\
        % \hline
        % Single model & 70.86 $\pm$ .36 &  9.8 $\pm$ .31 & 1.170 $\pm$ 0.0139 & 0.411 $\pm$ 0.004\\
        % Vanilla pooling \textbf{[A]} & 78.41 & 5.9 & 0.782 & 0.308\\
        % Pool-then-calibrate \textbf{[D]} & 78.41 & 5.1 & 0.779 & 0.307\\
        % \hline
        
        \multicolumn{5}{c}{Diabetic Retinopathy (5000 samples)}\\
        \hline
        Single model & 61.26 $\pm$ .62 &  2.96 $\pm$ .64 &  0.657 $\pm$ 0.004 &  0.465 $\pm$  0.004\\
        Vanilla pooling \textbf{[A]} & 64.38 & 4.9 & 0.641 & 0.450\\
        Pool-then-calibrate \textbf{[D]} & 64.38 & 2.9 $\pm$ .8 & 0.637 $\pm$ .002 & 0.446 $\pm$ .001\\
        \hline

        \multicolumn{5}{c}{Imagenette (1000 samples)}\\
        \hline
        Single model & 78.67 $\pm$ .34 & 14.45 $\pm$ .95 & 0.796 $\pm$ 0.012 &  0.332 $\pm$  0.005\\
        Vanilla pooling \textbf{[A]} & 80.91 &  18.2 &  0.753 &  0.312\\
        Pool-then-calibrate \textbf{[D]} & 80.91 & 3.5 $\pm$ 1.0 & 0.638 $\pm$ .005 & 0.273 $\pm$ .001\\
        \hline

        % \multicolumn{5}{c}{Imagewoof (1000 samples)}\\
        % \hline
        % Single model & 66.48 $\pm$ .62 & 7.31 $\pm$ .7 & 1.037 $\pm$ .013 & 0.464 $\pm$ .005\\
        % Vanilla pooling \textbf{[A]} & 66.89 & 8.9 & 1.044 & 0.452\\
        % Pool-then-calibrate \textbf{[D]} & 66.89 & 3.1 $\pm$ 2.1 & 1.045 $\pm$ .12 & 0.457 $\pm$ .003\\
        % \hline

        \multicolumn{5}{c}{MNIST (500 samples)}\\
        \hline
        Single model & 89.3 $\pm$ .8 & 6.4 $\pm$ .9 & 0.375 $\pm$ .022 & 0.163 $\pm$ .01\\
        Vanilla pooling \textbf{[A]} & 90.53 & 8.4 & 0.351 & 0.151\\
        Pool-then-calibrate \textbf{[D]} & 90.53 & 2.1 & 0.306 & 0.139\\
        \hline
    \end{tabular}
    \caption{Numerical result of Vanilla pooling \textbf{[A]} and Pool-then-Calibrate \textbf{[D]} for different setups. In our chosen setups, the pooled predictions are consistently more under-confident than single models. \textit{Pool-then-calibrate} has the best performance across all the metrics.}
    \label{tab:table_model_pool}
\end{table}

\paragraph{Detailed numerical results}
In table \ref{tab:table_model_pool} we present the detailed numerical results for all our setups. The table includes result of our proposed \textit{Pool-then-calibrate} method \textbf{[D]}, the vanilla pooling method \textbf{[A]}, and that of the individual models. The conclusions are consistent across all the setups. 

\bibliographystyle{plain}
\bibliography{uqdl}

\begin{thebibliography}{10}

\bibitem{ashukha2021pitfalls}
Arsenii Ashukha, Alexander Lyzhov, Dmitry Molchanov, and Dmitry Vetrov.
\newblock Pitfalls of in-domain uncertainty estimation and ensembling in deep
  learning, 2021.

\bibitem{bietti2019inductive}
Alberto Bietti and Julien Mairal.
\newblock On the inductive bias of neural tangent kernels.
\newblock In {\em Advances in Neural Information Processing Systems}, pages
  12873--12884, 2019.

\bibitem{VariationalBNNBlundel}
Charles Blundell, Julien Cornebise, Koray Kavukcuoglu, and Daan Wierstra.
\newblock Weight uncertainty in neural networks, 2015.

\bibitem{bonab2017less}
Hamed Bonab and Fazli Can.
\newblock Less is more: a comprehensive framework for the number of components
  of ensemble classifiers.
\newblock {\em arXiv preprint arXiv:1709.02925}, 2017.

\bibitem{brier1950verification}
Glenn~W Brier.
\newblock Verification of forecasts expressed in terms of probability.
\newblock {\em Monthly weather review}, 78(1):1--3, 1950.

\bibitem{chen2020simple}
Ting Chen, Simon Kornblith, Mohammad Norouzi, and Geoffrey Hinton.
\newblock A simple framework for contrastive learning of visual
  representations.
\newblock {\em arXiv preprint arXiv:2002.05709}, 2020.

\bibitem{eyepacs_kaggle_DR}
Jorge Cuadros and George Bresnick.
\newblock {EyePACS}: {An} {Adaptable} {Telemedicine} {System} for {Diabetic}
  {Retinopathy} {Screening}.
\newblock {\em Journal of Diabetes Science and Technology}, 3(3):509--516, May
  2009.

\bibitem{dziugaite2017computing}
Gintare~Karolina Dziugaite and Daniel~M Roy.
\newblock Computing nonvacuous generalization bounds for deep (stochastic)
  neural networks with many more parameters than training data.
\newblock {\em arXiv preprint arXiv:1703.11008}, 2017.

\bibitem{fort2019deep}
Stanislav Fort, Huiyi Hu, and Balaji Lakshminarayanan.
\newblock Deep ensembles: A loss landscape perspective.
\newblock {\em arXiv preprint arXiv:1912.02757}, 2019.

\bibitem{gabrie2018entropy}
Marylou Gabri{\'e}, Andre Manoel, Cl{\'e}ment Luneau, Nicolas Macris, Florent
  Krzakala, Lenka Zdeborov{\'a}, et~al.
\newblock Entropy and mutual information in models of deep neural networks.
\newblock In {\em Advances in Neural Information Processing Systems}, pages
  1821--1831, 2018.

\bibitem{MCD}
Y.~Gal and Z.~Ghahramani.
\newblock Dropout as a bayesian approximation.
\newblock {\em International Conference on Machine Learning}, 2016.

\bibitem{gal2016dropout}
Yarin Gal and Zoubin Ghahramani.
\newblock Dropout as a bayesian approximation: Representing model uncertainty
  in deep learning.
\newblock In {\em international conference on machine learning}, pages
  1050--1059, 2016.

\bibitem{gneiting2007strictly}
Tilmann Gneiting and Adrian~E Raftery.
\newblock Strictly proper scoring rules, prediction, and estimation.
\newblock {\em Journal of the American statistical Association},
  102(477):359--378, 2007.

\bibitem{PracVarInf}
A.~Graves.
\newblock Practical variational inference for neural networks.
\newblock {\em NIPS}, 2011.

\bibitem{graves2016hybrid}
Alex Graves, Greg Wayne, Malcolm Reynolds, Tim Harley, Ivo Danihelka, Agnieszka
  Grabska-Barwi{\'n}ska, Sergio~G{\'o}mez Colmenarejo, Edward Grefenstette,
  Tiago Ramalho, John Agapiou, et~al.
\newblock Hybrid computing using a neural network with dynamic external memory.
\newblock {\em Nature}, 538(7626):471--476, 2016.

\bibitem{Guo_calibration_2017}
C.~Guo, G.~Pleiss, Y.~Sun, and K.~Q. Weinberger.
\newblock On calibration of modern neural networks.
\newblock {\em Proceedings of the 34 th International Conference on Machine
  Learning, Sydney, Australia, PMLR 70, 2017}, 2017.

\bibitem{resnet_paper}
Kaiming He, Xiangyu Zhang, Shaoqing Ren, and Jian Sun.
\newblock Deep residual learning for image recognition.
\newblock {\em CoRR}, abs/1512.03385, 2015.

\bibitem{imagenette-woof}
Jeremy Howard.
\newblock {\em Imagenette and Imagewoof}, 2018.

\bibitem{jacot2018neural}
Arthur Jacot, Franck Gabriel, and Cl{\'e}ment Hongler.
\newblock Neural tangent kernel: Convergence and generalization in neural
  networks.
\newblock In {\em Advances in neural information processing systems}, pages
  8571--8580, 2018.

\bibitem{jose2008simple}
Victor Richmond~R Jose and Robert~L Winkler.
\newblock Simple robust averages of forecasts: Some empirical results.
\newblock {\em International journal of forecasting}, 24(1):163--169, 2008.

\bibitem{cifar10}
Alex Krizhevsky, Vinod Nair, and Geoffrey Hinton.
\newblock Cifar-10 (canadian institute for advanced research).
\newblock 2009.

\bibitem{DeepEnsemble}
B.~Lakshminarayanan, A.~Pritzel, and C.~Blundell.
\newblock Simple and scalable predictive uncertainty estimation using deep
  ensembles.
\newblock {\em 31st Conference on Neural Information Processing Systems, Long
  Beach, CA, USA}, 2017.

\bibitem{LeNet}
Y.~LeCun, B.~Boser, J.~S. Denker, D.~Henderson, R.~E. Howard, W.~Hubbard, and
  L.~D. Jackel.
\newblock {Backpropagation Applied to Handwritten Zip Code Recognition}.
\newblock {\em Neural Computation}, 1(4):541--551, 12 1989.

\bibitem{lecun-mnisthandwrittendigit-2010}
Yann LeCun and Corinna Cortes.
\newblock {MNIST} handwritten digit database.
\newblock 2010.

\bibitem{lee2015m}
Stefan Lee, Senthil Purushwalkam, Michael Cogswell, David Crandall, and Dhruv
  Batra.
\newblock Why m heads are better than one: Training a diverse ensemble of deep
  networks.
\newblock {\em arXiv preprint arXiv:1511.06314}, 2015.

\bibitem{uncertaintyForDiabRetin}
Christian Leibig, Vaneeda Allken, Murat~Se{\c{c}}kin Ayhan, Philipp Berens, and
  Siegfried Wahl.
\newblock Leveraging uncertainty information from deep neural networks for
  disease detection.
\newblock {\em Scientific Reports}, 7(1):17816, Dec 2017.

\bibitem{VariationDLMatGauss}
C.~Louizos and M.~Welling.
\newblock Structured and efficient variational deep learning with matrix
  gaussian posteriors.
\newblock {\em arXiv preprint arXiv:1603.04733}, 2016.

\bibitem{MacKay_1992b}
D.~J.~C. MacKay.
\newblock A practical bayesian framework for backpropagation networks.
\newblock {\em Neural Computation, 4(3):448–472}, 1992.

\bibitem{mackay2003information}
David~JC MacKay and David~JC Mac~Kay.
\newblock {\em Information theory, inference and learning algorithms}.
\newblock Cambridge university press, 2003.

\bibitem{SWAG}
W.~Maddox, T.~Garipov, P.~Izmailov, D.~Vetrov, and A.~G. Wilson.
\newblock A simple baseline for bayesian uncertainty in deep learning.
\newblock {\em arXiv preprint arXiv:1902.02476}, 2019.

\bibitem{mei2018mean}
Song Mei, Andrea Montanari, and Phan-Minh Nguyen.
\newblock A mean field view of the landscape of two-layer neural networks.
\newblock {\em Proceedings of the National Academy of Sciences},
  115(33):E7665--E7671, 2018.

\bibitem{mirowski2016learning}
Piotr Mirowski, Razvan Pascanu, Fabio Viola, Hubert Soyer, Andrew~J Ballard,
  Andrea Banino, Misha Denil, Ross Goroshin, Laurent Sifre, Koray Kavukcuoglu,
  et~al.
\newblock Learning to navigate in complex environments.
\newblock {\em arXiv preprint arXiv:1611.03673}, 2016.

\bibitem{mnih2015human}
Volodymyr Mnih, Koray Kavukcuoglu, David Silver, Andrei~A Rusu, Joel Veness,
  Marc~G Bellemare, Alex Graves, Martin Riedmiller, Andreas~K Fidjeland, Georg
  Ostrovski, et~al.
\newblock Human-level control through deep reinforcement learning.
\newblock {\em Nature}, 518(7540):529--533, 2015.

\bibitem{neal2012bayesian}
Radford~M Neal.
\newblock {\em Bayesian learning for neural networks}, volume 118.
\newblock Springer Science \& Business Media, 2012.

\bibitem{dataAugmentationEffectiveness}
Luis Perez and Jason Wang.
\newblock The effectiveness of data augmentation in image classification using
  deep learning.
\newblock {\em arXiv preprint arXiv:1712.04621}, 2017.

\bibitem{Platt}
J.~Platt.
\newblock Probabilistic outputs for support vector machines and comparisons to
  regularized likelihood methods.
\newblock {\em Advances in Large Margin Classifiers, 10(3)}, 1999.

\bibitem{prechelt1998early}
Lutz Prechelt.
\newblock Early stopping-but when?
\newblock In {\em Neural Networks: Tricks of the trade}, pages 55--69.
  Springer, 1998.

\bibitem{rezende2014stochastic}
Danilo~Jimenez Rezende, Shakir Mohamed, and Daan Wierstra.
\newblock Stochastic backpropagation and approximate inference in deep
  generative models.
\newblock {\em arXiv preprint arXiv:1401.4082}, 2014.

\bibitem{rotskoff2018neural}
Grant~M Rotskoff and Eric Vanden-Eijnden.
\newblock Neural networks as interacting particle systems: Asymptotic convexity
  of the loss landscape and universal scaling of the approximation error.
\newblock {\em arXiv preprint arXiv:1805.00915}, 2018.

\bibitem{srivastava2014dropout}
Nitish Srivastava, Geoffrey Hinton, Alex Krizhevsky, Ilya Sutskever, and Ruslan
  Salakhutdinov.
\newblock Dropout: a simple way to prevent neural networks from overfitting.
\newblock {\em The Journal of Machine Learning Research}, 15(1):1929--1958,
  2014.

\bibitem{szegedy2015going}
Christian Szegedy, Wei Liu, Yangqing Jia, Pierre Sermanet, Scott Reed, Dragomir
  Anguelov, Dumitru Erhan, Vincent Vanhoucke, and Andrew Rabinovich.
\newblock Going deeper with convolutions.
\newblock In {\em Proceedings of the IEEE conference on computer vision and
  pattern recognition}, pages 1--9, 2015.

\bibitem{subsequentBalaji}
Yeming Wen, Ghassen Jerfel, Rafael Muller, Michael~W. Dusenberry, Jasper Snoek,
  Balaji Lakshminarayanan, and Dustin Tran.
\newblock Combining ensembles and data augmentation can harm your calibration,
  2021.

\bibitem{wenzel2020good}
Florian Wenzel, Kevin Roth, Bastiaan~S Veeling, Jakub {\'S}wi{\k{a}}tkowski,
  Linh Tran, Stephan Mandt, Jasper Snoek, Tim Salimans, Rodolphe Jenatton, and
  Sebastian Nowozin.
\newblock How good is the bayes posterior in deep neural networks really?
\newblock {\em arXiv preprint arXiv:2002.02405}, 2020.

\bibitem{SVDKL}
Andrew~G Wilson, Zhiting Hu, Ruslan~R Salakhutdinov, and Eric~P Xing.
\newblock Stochastic variational deep kernel learning.
\newblock In {\em Advances in Neural Information Processing Systems}, pages
  2586--2594, 2016.

\bibitem{wilson2020bayesian}
Andrew~Gordon Wilson and Pavel Izmailov.
\newblock Bayesian deep learning and a probabilistic perspective of
  generalization.
\newblock {\em arXiv preprint arXiv:2002.08791}, 2020.

\bibitem{ProbForcastResearchPers}
Robert~L. Winkler, Yael Grushka-Cockayne, Kenneth~C. Lichtendahl, and Victor
  Richmond~R. Jose.
\newblock Probability forecasts and their combination: A research perspective.
\newblock {\em Decision Analysis}, 16(4):239--260, 2019.

\bibitem{subsequentXixin}
Xixin Wu and Mark Gales.
\newblock Should ensemble members be calibrated?, 2021.

\bibitem{generalizationDeepLearning}
Chiyuan Zhang, Samy Bengio, Moritz Hardt, Benjamin Recht, and Oriol Vinyals.
\newblock Understanding deep learning requires rethinking generalization.
\newblock {\em arXiv preprint arXiv:1611.03530}, 2016.

\bibitem{mixup_original}
Hongyi Zhang, Moustapha Ciss{\'{e}}, Yann~N. Dauphin, and David Lopez{-}Paz.
\newblock mixup: Beyond empirical risk minimization.
\newblock {\em CoRR}, abs/1710.09412, 2017.

\bibitem{regularizationElasticNet}
Hui Zou and Trevor Hastie.
\newblock Regularization and variable selection via the elastic net.
\newblock {\em Journal of the royal statistical society: series B (statistical
  methodology)}, 67(2):301--320, 2005.

\end{thebibliography}

\end{document}